\documentclass[conference]{IEEEtran}
\IEEEoverridecommandlockouts
\usepackage{cite}
\usepackage{amsmath,amssymb,amsfonts}
\usepackage{algorithmic}
\usepackage{graphicx}
\usepackage{textcomp}
\usepackage{amsmath,graphicx}
\usepackage{booktabs}
\usepackage{multirow}
\usepackage{xcolor}
\def\BibTeX{{\rm B\kern-.05em{\sc i\kern-.025em b}\kern-.08em
    T\kern-.1667em\lower.7ex\hbox{E}\kern-.125emX}}

\usepackage{fontawesome5}
\usepackage[colorlinks=true, linkcolor=black, urlcolor=black, citecolor=black,hidelinks]{hyperref}
\usepackage[capitalise]{cleveref}

\usepackage{xcolor}
\usepackage[normalem]{ulem}
\useunder{\uline}{\ulined}{}%
\DeclareUrlCommand{\bulurl}{}

\begin{document}

\title{AudioBERT: Audio Knowledge Augmented Language Model}

\author{
Hyunjong Ok$^{1,2\:\ast\dag}$\thanks{$^\dag$ corresponding author} \quad Suho Yoo$^3$$^\ast$\thanks{$^\ast$ equal contribution} \quad Jaeho Lee$^{1\dag}$ \\
$^1$POSTECH\qquad$^2$HJ AILAB\qquad$^3$Inha University\\
\texttt{\{hyunjong.ok, uso7d0\}@gmail.com, jaeho.lee@postech.ac.kr}
}

\maketitle

\begin{abstract}
Recent studies have identified that language models, pretrained on text-only datasets, often lack elementary visual knowledge, \textit{e.g.,} colors of everyday objects. Motivated by this observation, we ask whether a similar shortcoming exists in terms of the \textit{auditory} knowledge. To answer this question, we construct a new dataset called AuditoryBench, which consists of two novel tasks for evaluating auditory knowledge. Based on our analysis using the benchmark, we find that language models also suffer from a severe lack of auditory knowledge. To address this limitation, we propose AudioBERT, a novel method to augment the auditory knowledge of BERT through a retrieval-based approach. First, we detect auditory knowledge spans in prompts to query our retrieval model efficiently. Then, we inject audio knowledge into BERT and switch on low-rank adaptation for effective adaptation when audio knowledge is required. Our experiments demonstrate that AudioBERT is quite effective, achieving superior performance on the AuditoryBench. The dataset and code are available at  \bulurl{https://github.com/HJ-Ok/AudioBERT}.

%Recent studies have identified a lack of visual knowledge in language models as they trained on text-only data. Building upon this insight, we investigated whether language models keep auditory knowledge and found identical shortcomings. 
%To tackle this challenge, we propose \textbf{AudioBERT}, a novel method that injects auditory knowledge into BERT through a retrieval-based approach. By freezing BERT, AudioBERT utilizes auditory knowledge without decreasing the performance of the language model. 
%For evaluation, we introduce a new benchmark, \textbf{AuditoryBench}, which contains two novel tasks for evaluating auditory knowledge. One involves identifying animal sounds, and the other distinguishes between different pitches. Based on these two proposed tasks, we evaluate AudioBERT, demonstrating its superior performance.
\end{abstract}

\begin{IEEEkeywords}
Auditory Knowledge, Language Model, Retrieval Augmented Prediction
\end{IEEEkeywords}

\section{Introduction}
\label{sec:intro}

The advance of pretrained language models has spurred significant improvements across various language-related tasks \cite{devlin2019bert, radford2019language}, and has been extended to processing multimodal information \cite{llava,pengi}. However, a major limitation of popular language models is that they are pretrained only on textual data, which can lead to gaps in knowledge from other domains. Indeed, in visual domains, researchers have repeatedly found that common language models lack sufficient visual commonsense knowledge---such as the color of common objects---leading to a poor performance on visual tasks \cite{zhang2022visual, liu2022things, alper2023bert, rahmanzadehgervi2024vision}. In response, recent studies have proposed algorithms to augment language models with visual knowledge \cite{tan2020vokenization, lu2022imagination, wangvisually, tang2023learning}. 

\begin{figure}[t]
    \centering
    \includegraphics[width=0.7\columnwidth]{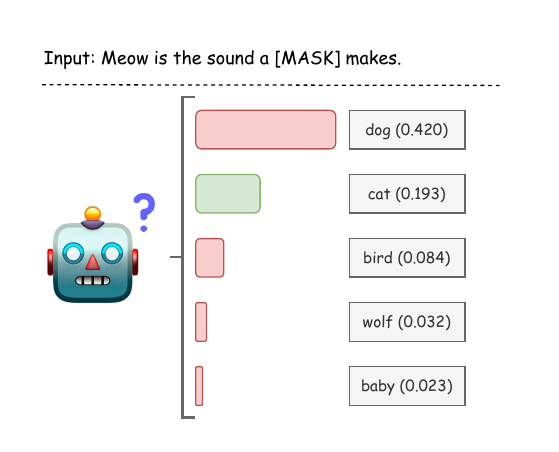}
    \caption{A visual illustration of the BERT prediction probabilities on a masked word, when asked to complete an example sentence from the AuditoryBench dataset which requires auditory commonsense knowledge.} \label{fig:our_task}
\end{figure}

Does the same limitation hold for \textit{auditory} commonsense knowledge? This question, unfortunately, has not been clearly addressed yet. Although a recent work studies the effectiveness of audio snippet embeddings in the language model representation space \cite{ngo2024language}, it is not known whether the language models have rich commonsense knowledge regarding the auditory signals, e.g., which animal make a specific sound (\cref{fig:our_task}).

To answer this question, we present \textbf{AuditoryBench}, the first benchmark dataset (to our knowledge) for evaluating the language models' auditory knowledge. In particular, we propose two auditory knowledge tasks: (1) animal sound recognition and (2) sound pitch comparison. The animal sound recognition task asks the language model to predict which animal is likely to make a sound that corresponds to specific echoing words (\textit{i.e.}, onomatopoeia), such as ``meow.'' The sound pitch comparison task asks the language model to predict which sound source (\textit{e.g.}, musical instruments, objects, or environments) is likelier to produce sound with a higher pitch. To construct this benchmark, we propose an LLM-based data-processing pipeline for the sake of the scalability of the benchmark dataset \cite{kim2023soda,mei2024wavcaps}.

Using AuditoryBench, we discover that language models severely lack auditory commonsense knowledge. In particular, we test three different language models---BERT\cite{devlin2019bert}, Gemma\cite{team2024gemma}, and LLaMA\cite{dubey2024llama}---and find that all models achieve low predictive accuracy in both benchmark tasks (\cref{tab:audio_knowledge_recognition}).

\begin{figure*}[t]
    \centering
    \includegraphics[width=0.85\linewidth]{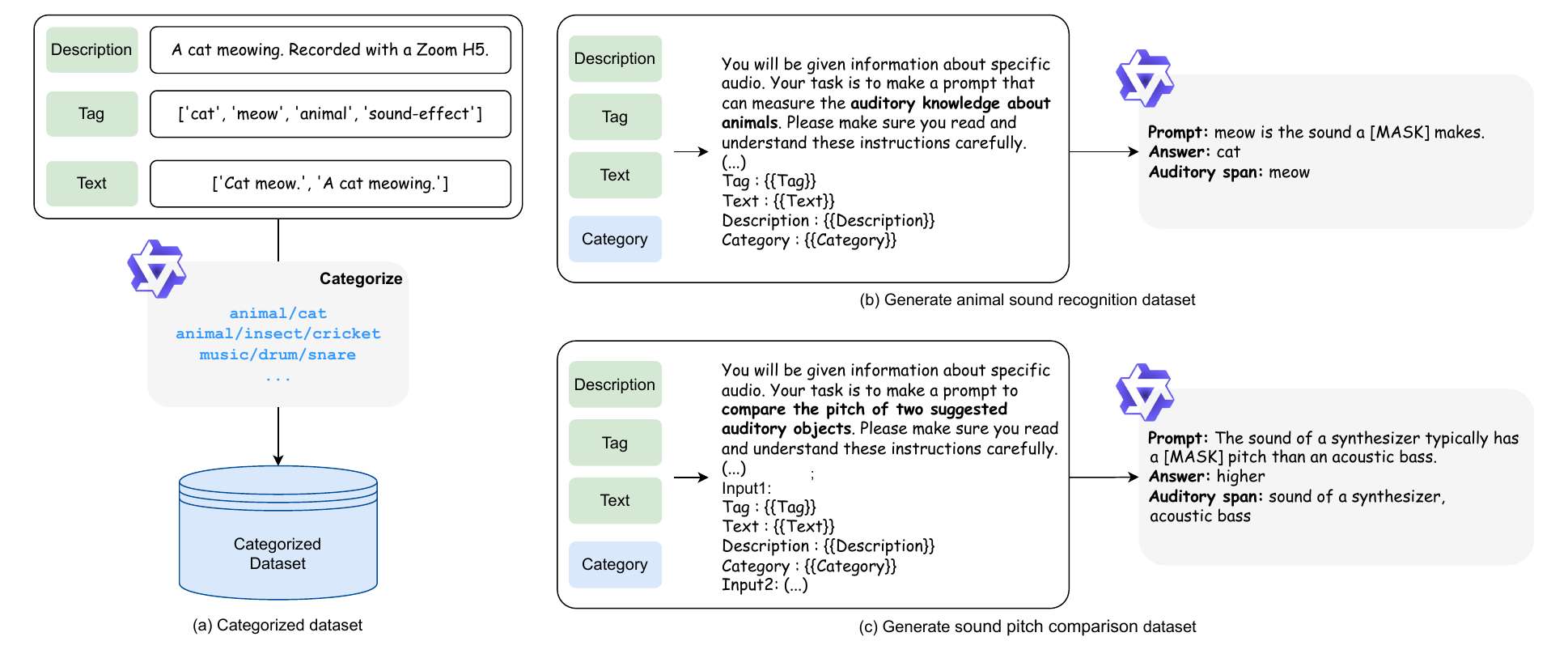}
    \caption{A visual illustration of the overall data generation process for the AuditoryBench.} \label{fig:datapipline_overview}
\end{figure*}

To address this shortcoming, we propose \textbf{AudioBERT}, a simple yet effective retrieval-based framework for injecting auditory knowledge into language models. Our approach involves detecting text spans where auditory knowledge is necessary. Whenever needed, relevant audio is retrieved by querying the detected text span to CLAP \cite{elizalde2023clap}, a model that measures the text-audio similarity. Then, the embedding of the retrieved audio sample is injected into the language model. Upon identification of auditory spans by the detector, the language model activates Low-Rank Adaptation (LoRA)\cite{hu2022lora} weights, which is finetuned with AudiotoryBench, which maintains its pretrained knowledge makes the model perform well in other tasks by deactivating LoRA weights. Our experiments demonstrate that the proposed method is quite effective, improving the prediction accuracy on AuditoryBench by more than 40\% in the test set. As far as we know, AudioBERT is the first algorithm to augment language models with auditory commonsense knowledge.

The key contributions of our work are threefold:
\begin{itemize}
\item We introduce the first benchmark dataset for evaluating the auditory knowledge of language models via an efficient LLM-based data generation pipeline.

\item We discover that common language models pretrained on text-only datasets suffer from a severe lack of auditory commonsense knowledge.

%\item We develop an LLM-based pipeline to efficiently create high-quality datasets for our benchmark tasks.

\item  We propose a novel and effective framework to inject auditory knowledge into pretrained language models, which adaptively retrieves and utilizes relevant auditory knowledge, utilizing switch-on-off LoRA to preserve pretrained knowledge.
    
\end{itemize}

\section{Methods}
\label{sec:methods}

\subsection{AuditoryBench}
To assess the auditory commonsense knowledge, we construct a dataset called AuditoryBench. AuditoryBench is based on LAION-Audio-630K \cite{wu2023large}, a large-scale audio-text dataset with over 600,000 audio-text pairs.

The first step to construct AuditoryBench is to categorize each audio sample of the dataset by processing the paired text information with an LLM (\cref{fig:datapipline_overview}(a)); the text information consists of the audio metadata (description and tag) and the paired text. Based on this information, we give a detailed instruction to categorize hierarchically (\textit{e.g.,} ``animal/dog/pomeranian'') with few-shot examples.

The next step is to construct tasks based on the generated categories and the text information. We design two tasks: Animal sound recognition and sound pitch comparison.

%To assess the auditory commonsense knowledge, we design two tasks and construct a dataset called AuditoryBench. The benchmark is based on the LAION-Audio-630K \cite{wu2023large}, a large-scale audio-text dataset (633,526 pairs) with over 4,000 hours of audio sourced from eight public websites. It includes various audio content, such as human activities, natural sounds, and audio effects, making it the largest publicly available audio-text dataset. First, as illustrated in Fig.\ref{fig:datapipline_overview}(a), we leveraged an LLM to generate categories for each audio sample based on the provided text descriptions. Then, we created the animal sound recognition task and sound pitch comparison task datasets using the Qwen2-72B-Instruct-AWQ \cite{yang2024qwen2} model. Detailed statistics are shown in Table \ref{tab:animal_stats}.

\begin{table}[t]
\caption{Statistics for animal sounds recognition and sound pitch comparison data.}
\centering
\vspace{0.25cm}
\resizebox{0.7\columnwidth}{!}{
\begin{tabular}{lccc}
\toprule
\textbf{Splits} & \textbf{\# Captions} & \textbf{\# Words/Caption} & \textbf{\# Total tokens}  \\ \midrule
\multicolumn{4}{c}{\textbf{Animal sound recognition}}\\ \midrule
Train & 4,211 & 9.27 & 39,024  \\ 
Dev & 593 & 9.39 & 5,567 \\ 
Test & 1,211 & 9.30 & 11,268  \\ 
Wiki & 197 & 7.01 & 1,381 \\ \midrule
Total & 6,212 & 9.21 & 57,240  \\ 
\midrule
%\textbf{Pitch} & \textbf{\# Captions} & \textbf{\# Words/Caption} & \textbf{\# Total tokens}  \\ \midrule
\multicolumn{4}{c}{\textbf{Sound pitch comparison}}\\ \midrule
Train & 8,312 & 18.18 & 151,081  \\ 
Dev & 1,178 & 18.12 & 21,343  \\ 
Test & 2,387 & 18.19 & 43,431  \\
Wiki & 3,625 & 12.41 & 44,987     \\ \midrule
Total & 15,502 & 16.83 & 260,842  \\ \bottomrule
\end{tabular}
}
\label{tab:animal_stats}
\end{table}

\begin{figure*}[t]
    \centering
    \includegraphics[width=0.85\linewidth]{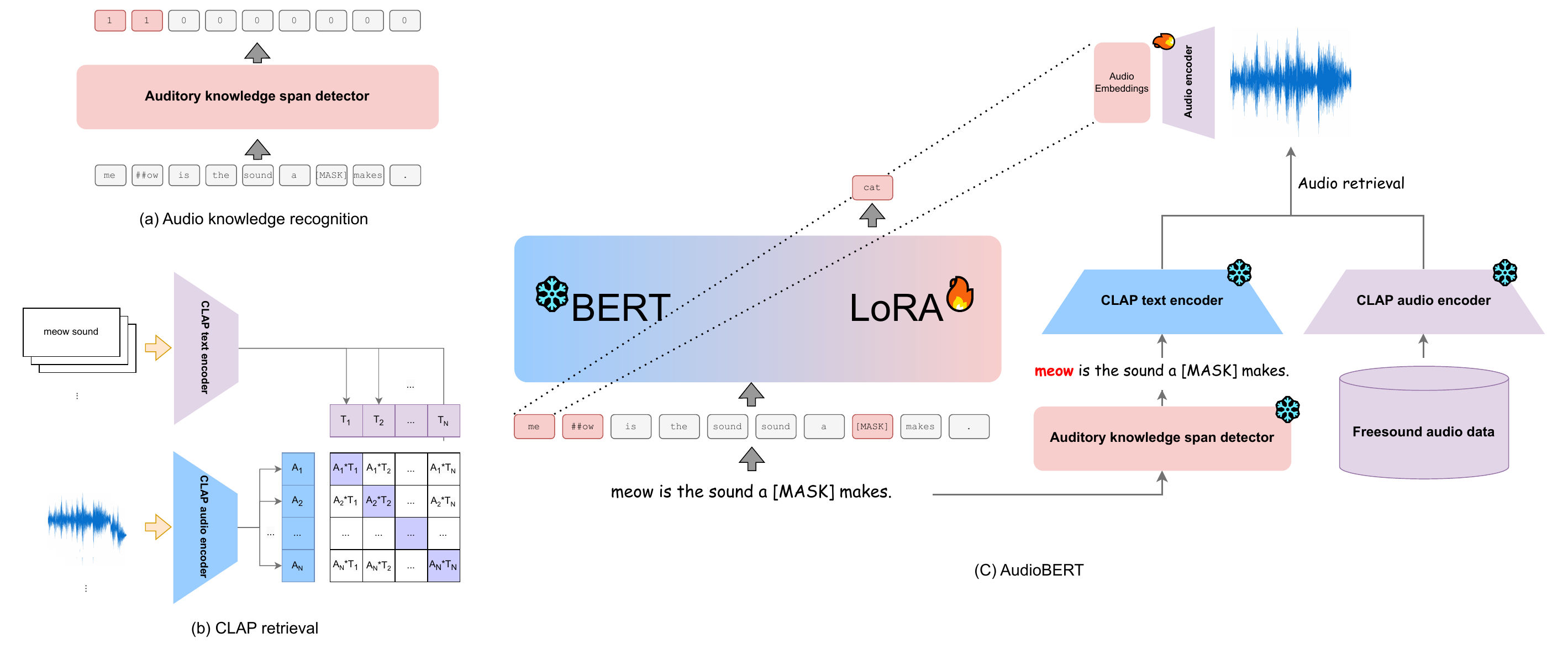}
    \caption{Overall process of our method} \label{fig:model_overview}
\end{figure*}

\paragraph*{Animal sound recognition}
This task asks to predict an animal from the given onomatopoeia that corresponds to an animal' sound. Each datum consists of a prompt (\textit{e.g.,} ``meow is the sound a [MASK] makes''), an answer (\textit{e.g.,} ``cat''), and an onomatopoeic span (\textit{e.g.,} ``meow''). This triplet is generated by first selecting a datum that is categorized as ``animal,'' and processing its text information with an LLM; see \cref{fig:datapipline_overview}(b) for an overall illustration and the prompt used. To enhance the data quality, we have additionally employed a human annotator to filter our inappropriate animal answers(e.g., “zombie, vampire”).

Using this approach, we have generated a task with 6,015 samples. The dataset is then split into training, development, and test sets with 70\%/10\%/20\% of samples. We have also collected animal sound data from Wikipedia,\footnote{\url{https://en.wikipedia.org/wiki/List_of_animal_sounds}} allocating them as additional test data. This additional set lets us assess the generalizability and quality of the dataset. %sec:data_quality  %({\color{red} pointer to results}).

%Animal sounds can be represented in text as onomatopoeia. As illustrated in Fig.\ref{fig:datapipline_overview}(b), we utilized a prompt-based process to generate datasets by input instructions into an LLM for data categorized as `animal.' For example, `meow is the sound a [MASK] makes' generated with the correct answer (in this case, `cat') and the onomatopoeic span (`meow'), which requires auditory knowledge. To reinforce data quality, we additionally employed a human annotator to filter out inappropriate labels.

%This approach generated a dataset comprising 6,015 samples. We partitioned the dataset into training, development, and test sets, allocating 70\%, 10\%, and 20\% of the samples, respectively. We additionally collected animal sounds from Wikipedia\footnote{\url{https://en.wikipedia.org/wiki/List_of_animal_sounds}}, utilizing this as a test set. This dataset lets us assess our dataset's generalization and quality.

\paragraph*{Sound pitch comparison}
This task asks to compare the pitch of two different sound sources. Each datum consists of a prompt (\textit{e.g.,} ``The sound of a synthesizer typically has a [MASK] pitch than an acoustic bass''), an answer (\textit{e.g.,} ``higher''), and two auditory spans (\textit{e.g.,} ``sound of a synthesizer,'' ``acoustic bass''). The triplet is generated by first selecting two audio samples, categorized into various music, objects, and environment categories (\cref{fig:datapipline_overview}(c)). Then, using the librosa library \cite{mcfee_2024_11192913}, we extract the average pitch for each audio sample. We then filter the data by selecting only the pairs in which the pitch difference exceeds 10\%, and an LLM can accurately classify which one is higher. 

We split the data into training, development, and test sets, with fractions 70\%/10\%/20\%. We have additionally collected the pitch range of musical instruments from Wikipedia\footnote{\url{https://en.wikipedia.org/wiki/Range_(music)}}, to utilize this as a test set. This dataset lets us assess our dataset’s generalization and quality.

%We can compare the pitch differences between the two audio samples that were given. As illustrated in Fig.\ref{fig:datapipline_overview}(c), we generate datasets based on the descriptions of the two audio samples using an LLM. We utilize data classified into music, objects, and environment categories, labeled by Fig.\ref{fig:datapipline_overview}(a) process. Subsequently, using the librosa \cite{mcfee_2024_11192913} library, we extract the average pitch values for each audio sample. We then filter the data by selecting only pairs where the pitch difference exceeds 10\% and where the LLM accurately classifies which one is higher. We partitioned the dataset into training, development, and test sets, allocating 70\%, 10\%, and 20\% of the samples, respectively. We additionally collected musical instruments pitch range from Wikipedia\footnote{\url{https://en.wikipedia.org/wiki/Range_(music)}}, utilizing this as a test set. This dataset lets us assess our dataset’s generalization and quality.

For both tasks, we use Qwen2-72B-Instruct-AWQ \cite{yang2024qwen2} as the LLM. Detailed statistics of these tasks are given in \cref{tab:animal_stats}.

\subsection{AudioBERT}
Now, we introduce our retrieval-based framework, which is coined AudioBERT. AudioBERT employs two models as its building block: An auditory knowledge span detector, which identifies the spans related to auditory knowledge within the given text, and CLAP \cite{elizalde2023clap}, a model that is used for retrieval. In what follows, we first describe each component (in \cref{subsec:method_1,subsec:method_2}, respectively) and then describe the overall AudioBERT framework in \cref{subsec:method_3}.

%First, we develop an auditory knowledge span detector to identify spans related to auditory knowledge within the text (Sec.\ref{subsec:method_1}). Then, we briefly describe CLAP \cite{elizalde2023clap}, a model we use for retrieval (Sec.\ref{subsec:method_2}). Finally, we introduce AudioBERT, a model that injects auditory knowledge into a frozen BERT model employing two preceding models (Sec.\ref{subsec:method_3}).

\subsubsection{Auditory knowledge span detector}
\label{subsec:method_1}
This model extracts the audio-relevant span from the given text, which can work as a good query. To obtain such model, we simply train a transformer encoder \cite{NIPS2017_3f5ee243} to classify the auditory knowledge span from other tokens, using the cross-entropy loss (\cref{fig:model_overview}(a)).

%Our approach involves a retrieval process to acquire relevant audio knowledge. For this process, we extract audio spans from the given text to retrieve with an optimized query. To train the audio span extraction model, we progress through a token classification method employing a transformer encoder\cite{NIPS2017_3f5ee243}, where the model is trained to perform binary classification to identify the audio knowledge spans. The classification process is optimized utilizing a cross-entropy loss.

% \begin{table}[t]
% \caption{.}
% \centering
% \vspace{0.25cm}
% \resizebox{1\linewidth}{!}{%
% \begin{tabular}{lcccccc}
% \toprule
% \multirow{2}{*}{\textbf{Trained w/}} & \multicolumn{3}{|c|}{\textbf{Animal sound}}  &\multicolumn{2}{|c}{\textbf{Height of sound}}  \\ 
% & \multicolumn{1}{|c}{Dev.}  & Test & Wiki Test & \multicolumn{1}{|c}{Dev.} &Test \\ \hline
% Animal sound & 92.35 $\pm$ 0.15 & 92.53 $\pm$ 0.25 & 100 $\pm$ 0.00 & 34.88 $\pm$ 4.09 & 34.51 $\pm$ 4.27 \\
% Height of sound & 51.77 $\pm$ 6.56 & 51.50 $\pm$ 6.19 & 94.76 $\pm$ 0.13 & 95.04 $\pm$ 0.09 & 79.08 $\pm$ 13.75 \\
% Combined dataset & 91.59 $\pm$ 0.28 & 92.41 $\pm$ 0.35 & 100 $\pm$ 0.00 & 94.73 $\pm$ 0.28 & 95.08 $\pm$ 0.30 \\
% \hline
% \end{tabular}
% }
% \label{tab:audio_knowledge_recognition}
% \end{table} 

\begin{table*}[!]
\caption{Experiment results of audio knowledge recognition using F1-score metrics.}
\centering
\vspace{0.25cm}
\resizebox{1\textwidth}{!}{%
\begin{tabular}{lcccccccccc}
\toprule
\multirow{2}{*}{\textbf{Trained w/}} & \multicolumn{3}{|c|}{\textbf{Animal sound}}  &\multicolumn{3}{|c}{\textbf{Sound pitch comparison}} &\multicolumn{3}{|c}{\textbf{Combined}} \\ 
& \multicolumn{1}{|c}{Dev.}  & Test & Wiki Test & \multicolumn{1}{|c}{Dev.} &Test & Wiki Test & \multicolumn{1}{|c}{Dev.} &Test & Wiki Test \\ \midrule
Animal sound & 92.75 $\pm$ 0.37 & 92.76 $\pm$ 0.19 & 100 $\pm$ 0.00 & 34.87 $\pm$ 8.48 & 34.61 $\pm$ 8.17 & 60.37 $\pm$ 12.61 & 43.67 $\pm$ 7.01 & 43.72 $\pm$ 6.77 & 60.96 $\pm$ 12.42\\
Height of sound & 52.63 $\pm$ 6.73 & 52.19 $\pm$ 6.40 & 79.08 $\pm$ 13.75 & 94.76 $\pm$ 0.13 & 95.04 $\pm$ 0.09 & 91.12 $\pm$ 2.12  & 88.64 $\pm$ 0.47 & 95.04 $\pm$ 0.43 & 89.89 $\pm$ 2.25\\
Combined dataset & 92.00 $\pm$ 0.39 & 92.67 $\pm$ 0.25 & 100 $\pm$ 0.00 & 94.88 $\pm$ 0.24 & 95.12 $\pm$ 0.24 & 92.47 $\pm$ 1.76 & 94.55 $\pm$ 0.19 & 94.84 $\pm$ 0.21 & 92.06 $\pm$ 1.74\\
\bottomrule
\end{tabular}
}
\label{tab:audio_knowledge_recognition}
\end{table*}

\begin{table*}[!]
\caption{Experiment results of AudioBERT using accuracy metrics.}
\centering
\vspace{0.25cm}
\resizebox{0.94\textwidth}{!}{%
\begin{tabular}{lcccccccccc}
\toprule
\multirow{2}{*}{\textbf{Methods}} & \multicolumn{3}{|c|}{\textbf{Animal sound}}  &\multicolumn{3}{|c}{\textbf{Sound pitch comparison}} &\multicolumn{3}{|c}{\textbf{Combined}} \\ 
& \multicolumn{1}{|c}{Dev.}  & Test & Wiki Test & \multicolumn{1}{|c}{Dev.} &Test & Wiki Test & \multicolumn{1}{|c}{Dev.} &Test & Wiki Test \\ \midrule
BERT-base\cite{devlin2019bert} & 15.51 & 13.46 & 3.05 & 59.42 & 60.41 & 48.06 & 44.72 & 44.61 & 45.72\\
BERT-large\cite{devlin2019bert} & 16.53 & 15.85 & 5.58 & 59.59 & 58.90 & 54.90 & 45.17 & 44.41 & 52.36\\
RoBERTa-base\cite{liu2019roberta} & 14.67 & 14.04 & 2.54 & 54.50 & 55.84 & 47.45 & 41.16 & 41.77 & 45.14\\
RoBERTa-large\cite{liu2019roberta} & 16.36 & 14.70 & 7.61 & 55.26 & 56.64 & 48.80 & 42.23 & 42.52 & 46.68\\
Gemma2-2B\cite{team2024gemma} & 14.33 & 15.11 & 6.60 & 59.25 & 60.45 & 47.86 & 44.21 & 45.19 & 45.73\\
LLaMA3.1-8B\cite{dubey2024llama} & 23.10 & 21.80 & \textbf{16.24} & 61.46 & 62.55 & 47.72 & 48.62 & 48.83 & 46.10\\
\midrule
Ours (BERT-base) & \textbf{38.28} $\pm$ 0.27 & 36.63 $\pm$ 0.38 & 14.32 $\pm$ 1.66 & 73.18 $\pm$ 0.95 & 74.83 $\pm$ 0.70 & 55.31 $\pm$ 2.38 & 61.49 $\pm$ 0.72 & 61.97 $\pm$ 0.59 & 53.20 $\pm$ 2.34 \\
Ours (BERT-large) & 37.30 $\pm$ 0.47 & \textbf{36.69} $\pm$ 0.27 & 15.03 $\pm$ 1.17 & \textbf{75.11} $\pm$ 0.80 & \textbf{76.31} $\pm$ 0.60 & \textbf{56.27} $\pm$ 1.18 & \textbf{62.45} $\pm$ 0.69 & \textbf{62.97} $\pm$ 0.49 & \textbf{54.14} $\pm$ 1.18 \\

\bottomrule
\end{tabular}
}
\label{tab:audio_knowledge_recognition2}
\end{table*} 

% We use BERT for mask token prediction for inference auditory task & Token classification for auditory knowledge span detect that used to autio retrieval.

\subsubsection{CLAP: Audio-text contrastive learning}
\label{subsec:method_2}
To retrieve the relevant audio from the detected auditory knowledge span, we utilize the audio-text model called CLAP \cite{elizalde2023clap}. CLAP is trained via audio-text contrastive learning: Given a batch of N audio-text pairs, the cosine similarity between the representations of matched pairs (positive) is maximized, while the cosine similarity is minimized in unmatched (negative) pairs. The loss function is formalized as follows:

%We utilized the audio-text model CLAP \cite{elizalde2023clap} to retrieve related audio for the auditory knowledge span. CLAP utilized contrastive learning in the pretrain phase. Given a batch of N audio-text pairs, the cosine similarity between the representations of matched pairs (positive) is maximized, while the cosine similarity is minimized in unmatched (negative) pairs. The loss function is formalized as follows:

{\fontsize{9}{12}\selectfont
\begin{equation}
L_\text{Audio} = - \frac{1}{N} \sum_{i=1}^N \log \frac{\exp(\text{cos}(A_i, T_i) / \tau)}{\sum_{j=1}^N \exp(\text{cos}(A_i, T_j) / \tau)}
\end{equation}

\begin{equation}
L_\text{Text} = - \frac{1}{N} \sum_{i=1}^N \log \frac{\exp(\text{cos}(A_i, T_i) / \tau)}{\sum_{j=1}^N \exp(\text{cos}(A_j, T_i) / \tau)}
\end{equation}

\begin{equation}
L = \frac{1}{2}( L_\text{Audio} + L_\text{Text} )
\end{equation}
}

$A_k$ and $T_k$ are the audio and text embeddings, $\text{cos}(\cdot,\cdot)$ denotes cosine similarity, and $\tau$ is a temperature parameter. 

%In the absence of ground truth audio labels, we collect audio samples by querying YouTube with the format `{Auditory span} sound of {Answer}' for animal sound recognition. We compute similarity against our collected audio samples from the LAION-Audio-630K, and the audio with the highest similarity is selected as the ground truth. For the sound pitch comparison task, we utilized the audio samples initially labeled.

\subsubsection{AudioBERT}
\label{subsec:method_3}
AudioBERT injects auditory knowledge into a language model using the auditory knowledge span detector and CLAP. Our approach works in three steps: We retrieve the relevant audio from the database using the span detector and the text encoder of CLAP, generate embedding using the CLAP audio encoder, and add this embedding to the first token of the auditory knowledge span (\cref{fig:model_overview}(c)).
During the training, we employ masked language modeling loss and LoRA to finetune the language model's LoRA weights and the audio encoder while keeping the other parameters free. For inference, the auditory knowledge span detector first determines whether auditory knowledge is required. If not, inference proceeds without activating the LoRA weights, maintaining the model's original pretrained weights. However, for tasks that require auditory knowledge, the auditory knowledge span detector recognizes this requirement, retrieves relevant audio, activates the LoRA weights, and injects the audio embedding into the model.
This novel architecture enables AudioBERT to dynamically adapt to tasks requiring auditory knowledge while maintaining its performance on general language understanding tasks.

\section{Experiments}
\paragraph*{Evaluation metrics}
We employ accuracy and F1-score for our metrics. Higher scores indicate better performance.

\paragraph*{Implementation details} 
In our implementation, we employ a BERT-base model for the auditory knowledge span detector. We trained with \(5\) epochs with a batch size of \(16\), a learning rate of \(1 \times 10^{-5}\), and utilizing AdamW\cite{loshchilov2018decoupled} optimizer. 

For AudioBERT training, we experimented using BERT for the language model and employed an AST\cite{gong21b_interspeech} encoder for auditory knowledge embedding injecting. We trained with \(20\) epochs with a batch size of \(32\), a learning rate of \(3 \times 10^{-4}\), and utilizing AdamW optimizer. For LoRA, we set the rank and alpha to 64 and 128, respectively.

We report the average score from 5 runs with different random seeds for each setting, utilizing Intel Gaudi 2 and NVIDIA H100.

\section{Results}

\subsection{Experiments results}
\subsubsection{Auditory knowledge span detector}
We report our auditory knowledge span detector performance in various trained data in Table \ref{tab:audio_knowledge_recognition}. The results suggest that training on single data may perform poorly, but the combined dataset works even with performance gain in sound pitch comparison.
\subsubsection{AudioBERT}
To evaluate the effectiveness of our approach, we compared our model against the existing language models: BERT, RoBERTa, Gemma2-2B, and LlaMA3.1-8B. As shown in Table \ref{tab:audio_knowledge_recognition2}, language models show the absence of auditory knowledge, and AudioBERT exhibits competitive performance, demonstrating its effectiveness in augmenting auditory knowledge.

% \subsection{Ablation study}

% \subsection{Effect of our methods}
% %lora train 전후
% %CLAP finetune 전후 비교

\subsection{Data quality assessment}
\label{sec:data_quality}
%LLM은 data generation에서 효과적이지만 우리 task는 LM이 성공적으로 수행하지 못하는 태스크로 잘못 생성할 수 있다. 이를 위해 우리는 animal sound를 사람이( thanks to 수호) 직접 필터링하는 과정을 거쳤고, height of sound의 경우 실제 pitch를 통해 필터링하는 과정을 거쳤다. 이 뿐만 아니라 우리는 wiki에서 직접 정답이 있는 추가적인 데이터셋을 test set에 추가적으로 넣었고, 세 명의 휴먼 이벨을 통해서 잘 생성했는지 물어보았다. (표 보여주기) 
%wiki test가 우리 데이터의 퀄리티 검증을 도와주고 height of sound의 경우 필터링 (height 10%이상 데이터 쓰기 전/후) 전/후 성능을 비교해서 퀄리티 보여주기
We took the following steps to ensure the quality and robustness of our data: When constructing the AuditoryBench, we first filtered the data. For animal sound recognition, human annotators filtered out inappropriate labels. For the sound pitch comparison task, we measured the pitch of each audio sample and maintained those correctly labeled. Additionally, we augmented the test set with information from Wikipedia to measure generalization and robustness. Finally, we conducted human evaluations on the test set for each benchmark. Three human annotators evaluated the criteria of accuracy and fluency, scoring out of 100 and 3, respectively. Accuracy reflects whether the labels were correctly applied, while fluency measures grammar, spelling, punctuation, word choice, and sentence structure. As shown in Table \ref{tab:human evaluation}, AuditoryBench exhibited high quality and robustness.

\begin{table}[t]
\caption{The result of human evaluation on the AuditoryBench dataset. We randomly sampled 100 data from each benchmark and asked the participants to rank the outputs of different systems' accuracy and fluency.
} 
\centering
\vspace{0.25cm}
\resizebox{0.7\columnwidth}{!}{%
\begin{tabular}{lcc}
\toprule
 Benchmark & Acc. & Fluency   \\ \midrule
Animal sound recognition & 92.00 & 2.91 \\
Sound pitch comparison & 87.00 & 3.00  \\
\bottomrule
\end{tabular}%
}
\label{tab:human evaluation}
\end{table} 

\section{Conclusions}
We introduce AuditoryBench, the first benchmark to assess the auditory knowledge of language models and propose AudioBERT, a novel method for injecting auditory knowledge into language models. We believe that our discovery of the lack of auditory knowledge in LMs can have implications for audio-language multimodal research and contribute to the research of language models effectively adapting to diverse modalities.

\section*{Acknowledgments}
This work was supported in part by the artificial intelligence industrial convergence cluster development project funded by the Ministry of Science and ICT (MSIT, Korea) \& Gwangju Metropolitan City, and in part by the National Research
Foundation of Korea (NRF) grant funded by the Korea government (MSIT) (No. RS-2024-00453301).

\vfill\pagebreak

\bibliographystyle{IEEEbib}
\bibliography{strings,refs}

\end{document}